\documentclass[11pt]{article}
\usepackage[final]{acl}
\usepackage{times}
\usepackage{latexsym}
\usepackage[T1]{fontenc}
\usepackage[utf8]{inputenc}

\usepackage{microtype}
\usepackage{inconsolata}

\usepackage{graphicx}
\usepackage{amsmath}
\usepackage{amssymb}
\usepackage{mathtools}
\usepackage{amsthm}
\usepackage{bm}
\usepackage{mathrsfs}
\usepackage{pifont}
\usepackage{array}
\usepackage{booktabs}
\usepackage{float}
\usepackage{tabularx}

\usepackage[table,dvipsnames]{xcolor}

\usepackage[linesnumbered,ruled,vlined]{algorithm2e}

\hypersetup{
    pagebackref=false,
    breaklinks=true,
    colorlinks=true,
    bookmarks=false,
    citecolor=blue,
    linkcolor=blue
}

\usepackage{amsfonts}

\title{SpatialGuard: Harness-Guided Verifiable Spatial Reasoning for Text-to-Image Generation}

\author{
 \textbf{Ziyun Qian\textsuperscript{1,2}},
 \textbf{Zizhi Chen\textsuperscript{1,2}},
 \textbf{Yizhou Liu\textsuperscript{1,2}},
 \textbf{Mingyang Sun\textsuperscript{1,2}},
\\
 \textbf{Dingkang Yang\textsuperscript{1,2,*$\,\dagger$}},
 \textbf{Lihua Zhang\textsuperscript{1,2,*}}
\\
\\
 \textsuperscript{1}College of Intelligent Robotics and Advanced Manufacturing, Fudan University
\\
 \textsuperscript{2}Fysics Intelligence Technologies Co., Ltd. (Fysics AI)
}

\begin{document}
\maketitle
\begingroup
\renewcommand{\thefootnote}{*}
\footnotetext{Corresponding authors. $\,^{\dagger}$Project lead.}
\endgroup
\begin{abstract}
Complex 3D spatial text to image generation requires models to convert natural language into stable visual geometry, not merely semantic appearance. Existing prompt-driven or layout-conditioned methods improve controllability, but often lack an optimizable and verifiable spatial intermediary before visual sampling. As a result, object relations, occlusion, visibility, and camera constraints can decay during multi-round generation. This paper presents SpatialGuard, a structured layout-guided framework for complex 3D spatial text-to-image generation. SpatialGuard parses prompts into image synthesis-oriented 3D layouts through a Spatial Layout Architect, realizes them as visual conditions and candidate images through a Visual Realizer, and uses a Visual Alignment Critic to validate consistency among prompt, layout, and image. To keep constraints stable across iterations, SpatialGuard introduces a Layout Harness that organizes rule constraints, tool invocation, shared knowledge, and feedback loops around the editable layout state. This design turns complex spatial generation from implicit prompt following into a verifiable process of planning, realization, validation, and repair. Comprehensive experiments show that SpatialGuard achieves state-of-the-art performance in complex 3D spatial layout generation and improves spatial faithfulness over existing text-to-image and layout control baselines.
\end{abstract}

\section{Introduction}

Complex 3D spatial text-to-image generation is a key capability for bridging natural-language interaction and controllable visual content production, with important applications in game-asset previewing \cite{game1,game2}, film storyboard design \cite{film1,film2}, and embodied-intelligence simulation \cite{embodied1}. These applications require models to translate spatial intent in natural language into stable visual geometry, rather than merely generating semantically related appearance content. Traditional prompt-driven methods \cite{Janus-pro-r1, Self-cross, NextStep-1, T2i-r1} mainly rely on text-to-image models to infer spatial structure from text implicitly. Thus, in relation-dense, view-sensitive, and layout-constrained scenes, they often satisfy only local semantics and struggle to maintain global spatial consistency. Recent methods \cite{Crea, SceneDesigner,st3d, Layercraft} introduce layout conditions, hierarchical generation, 3D control signals, and multi-step planning to make spatial structure more explicit during generation, significantly improving the controllability of complex scenes.
However, existing methods \cite{Compass_control,li2025mccd} still face two key limitations. First, the generation process usually lacks an image-synthesis-oriented spatial layout intermediary that is optimizable and verifiable. As a result, complex spatial intent lacks a stable carrier before visual sampling, and text alone fails to provide sufficiently explicit and executable spatial constraints for subsequent synthesis. Second, when complex spatial intent requires multiple rounds of planning, generation, evaluation, and correction, large models are prone to context forgetting and constraint decay. Spatial requirements established in early stages are difficult to preserve in later interactions, weakening fidelity to the original prompt and degrading generation quality.

\begin{figure*}[t]
  \centering
  \includegraphics[width=1.0\linewidth]{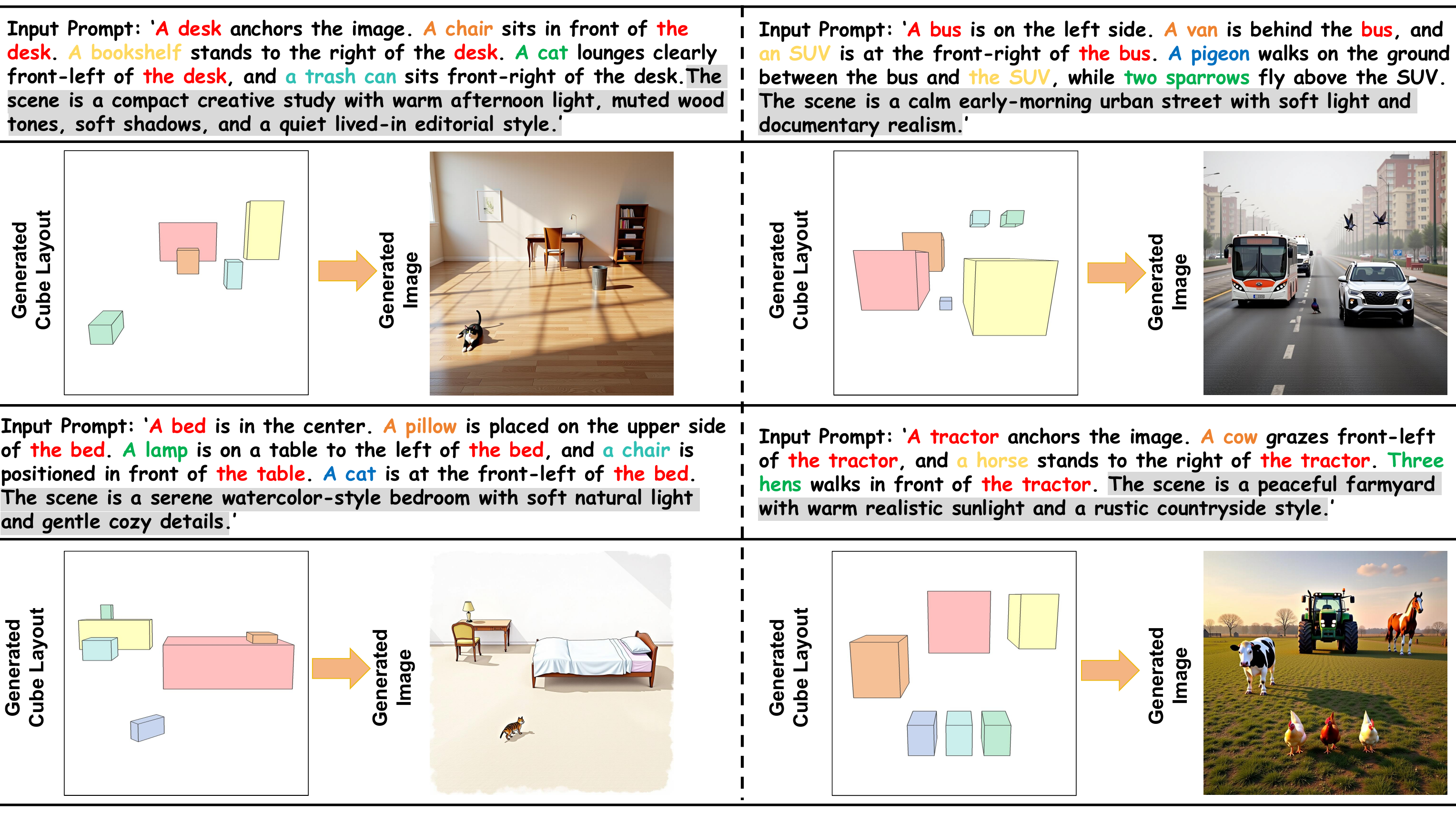}
  \caption{Example results of SpatialGuard on complex spatial text-to-image generation. Each case presents the input prompt, the generated 3D cube layout, and the corresponding synthesized image across urban street, farmyard, study, and bedroom scenes, demonstrating the model’s ability to preserve multi-object spatial relations, directional constraints, and scene-style descriptions.
  }
\label{fig:framework}
\end{figure*}

These limitations highlight two key requirements for complex spatial text-to-image generation. First, the model needs to form a verifiable layout representation before image sampling, allowing spatial constraints in text to be explicitly encoded, continuously optimized, and checked before generation. Second, the system needs a cross-round structured constraint enforcement mechanism, so that spatial requirements do not remain merely in prompts or scoring signals but are stably preserved, actively retrieved, and updated according to feedback across planning, execution, verification, and revision.
The former provides explicit spatial targets for image synthesis, while the latter ensures that these targets remain effective in long-chain interactions despite memory decay. Together, they indicate a stronger generation paradigm: complex 3D layouts should not only be prompted by language, but also explicitly designed, executed, and maintained.

To address these issues, we propose SpatialGuard, a structured-layout-guided generation framework for complex 3D spatial text-to-image generation. It unifies verifiable layout representation and persistent constraint enforcement within a single generation process. SpatialGuard consists of three collaborative modules. The Spatial Layout Architect parses input text into an image-synthesis-oriented 3D structured layout, providing a checkable and editable representation of complex spatial intent before sampling. The Visual Realizer converts this layout into visual conditions and synthesizes images, bringing spatial targets into pixel space. The Visual Alignment Critic evaluates consistency among text, layout, and image, and feeds deviations back to subsequent planning. These modules operate around a shared layout state, preventing text understanding, image realization, and result evaluation from becoming decoupled across multi-round interaction.

Moreover, to prevent memory decay of spatial constraints during multi-round agent planning and interaction, we first introduce the idea of an agent harness \cite{harness1,harness_survey,harness2} into text-to-image generation and distill it into four core mechanisms: rule constraints, tool invocation, shared knowledge, and feedback loops. Rule constraints convert spatial requirements in language into checkable layout boundaries. Tool invocation turns layout decisions into externally executable operations. Shared knowledge preserves layout versions, evaluation judgments, and revision trajectories across stages. Feedback loops reinject detected spatial deviations into layout planning and image generation. Based on the automated project workflow, SpatialGuard generates an initial 3D layout from text, performs deterministic verification according to the object list and spatial relations, iteratively invokes 3D layout rendering and image synthesis, and produces executable repair plans when spatial violations are detected. Thus, it shifts complex spatial text-to-image generation from prompt-driven implicit generation to verifiable generation guided by structured layouts and maintained by constraint mechanisms.

The main contributions of this paper are as follows:
\begin{itemize}
\item We propose SpatialGuard, an agentic layout generation framework for complex 3D spatial text-to-image generation, enabling closed-loop modeling from textual spatial intent to verifiable visual generation.
\item We introduce the concept of agent harness into the text-to-image pipeline for the first time and build a cross-round spatial constraint enforcement mechanism to mitigate memory decay in complex spatial generation by large models.
\item Comprehensive quantitative and qualitative experiments show that SpatialGuard achieves new state-of-the-art performance in complex 3D spatial text-to-image generation and significantly outperforms existing text-to-image and layout-control baselines.
\end{itemize}

\begin{figure*}[t]
  \centering
  \includegraphics[width=1.0\linewidth]{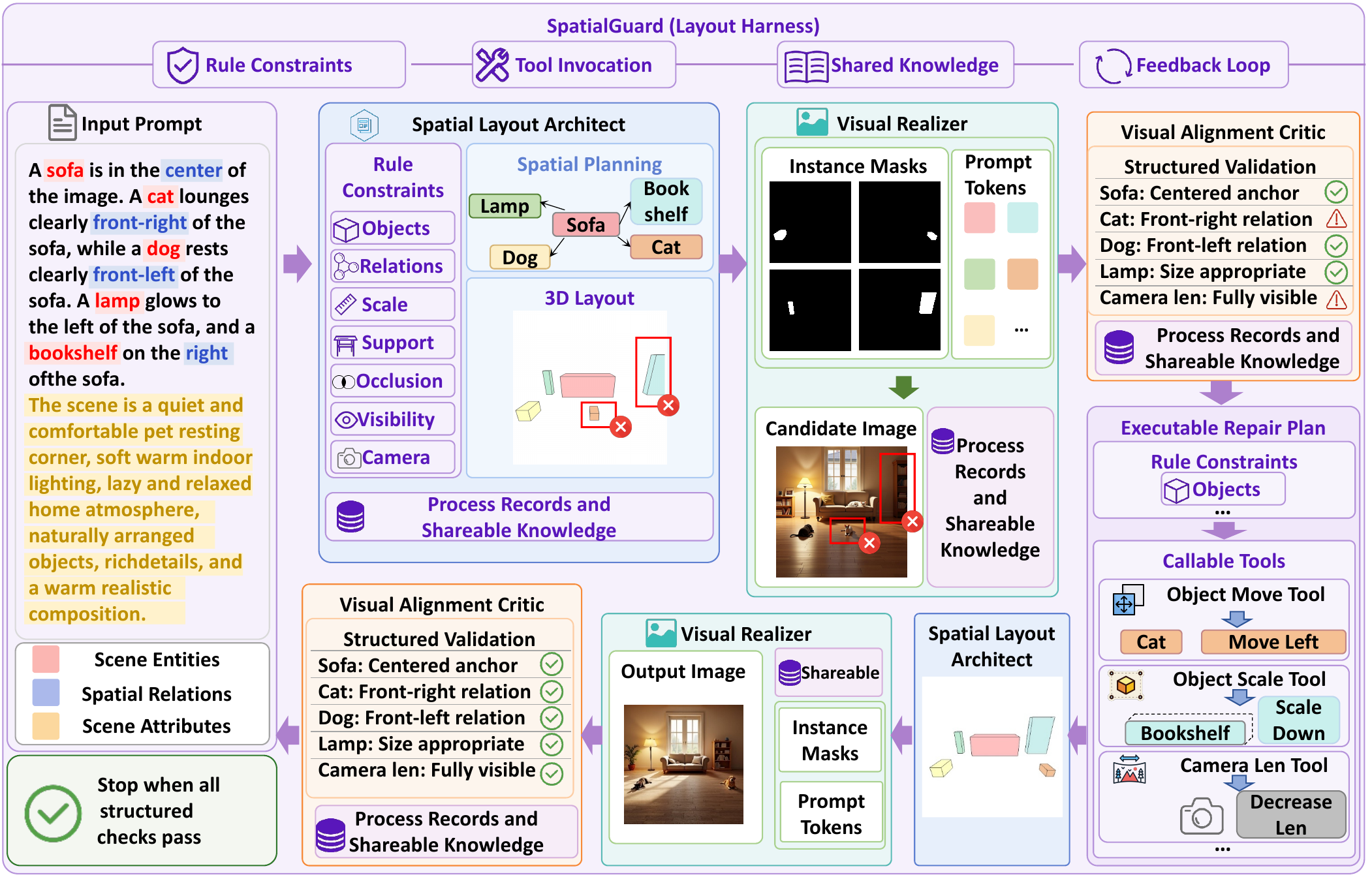}
  \caption{Overview of SpatialGuard. Given an input prompt, the Spatial Layout Architect parses objects, spatial relations, and scene appearance into a verifiable 3D layout under the Layout Harness. The Visual Realizer converts the layout into masks, token-aligned conditions, and a candidate image. The Visual Alignment Critic performs structured validation over object placement, relations, visibility, and camera constraints. Failed constraints are converted into executable repair actions and fed back to update the layout, while rule constraints, tool invocation, shared knowledge, and feedback loops preserve spatial intent across iterations until the final image satisfies the prompt.
  }
\label{fig:framework}
\end{figure*}

\section{Related Work}
\label{sec:related_work}

\subsection{Text to Image Generation}
\label{sec:related_t2i}

Text-to-image generation has progressed from high-fidelity synthesis to more deliberate modeling of semantic and spatial intent. General models improve prompt adherence through reasoning, reward optimization, autoregressive modeling, and spatial evaluation signals \cite{yang1,yang2,jiang1,jiang2,jiang3}. Janus-Pro-R1 \cite{Janus-pro-r1} and T2I-R1 \cite{T2i-r1} connect visual comprehension with generation through reinforcement learning and chain of thought style planning, while NextStep-1 \cite{NextStep-1} explores scalable autoregressive generation with continuous visual tokens \cite{NextStep-1}. Self-Cross \cite{Self-cross} reduces subject mixing among similar objects, and SpatialScore provides a reward model for spatial relation evaluation and reinforcement learning \cite{image1}. Recent benchmarks show that compositional and spatial alignment have become central axes for evaluating text-to-image models \cite{Genspace, SpatialGenEval, T2i-compbench++}. These works strengthen semantic alignment, but spatial composition is still largely learned or rewarded through implicit model behavior. A complementary line of work makes spatial control more explicit through planning, layout, pose, and 3D conditions. LayoutGPT \cite{Layoutgpt} uses language models for compositional visual planning, while MCCD \cite{li2025mccd} and CREA \cite{Crea} explore multi-agent collaboration for complex composition and creative generation. LayerCraft \cite{Layercraft} decomposes generation into layered reasoning and object integration. Compass-Control \cite{Compass_control} focuses on object orientation in multi-object scenes, SceneDesigner \cite{SceneDesigner} models 9 DoF object pose control, and SeeThrough3D \cite{st3d} introduces occlusion-aware 3D layout conditioning with camera control. These studies show that layouts, poses, layers, and planning improve controllability. However, existing methods still lack an optimizable and verifiable spatial layout intermediary, leaving complex spatial intent without a stable carrier before visual sampling and weakening executable spatial constraints for synthesis.

\subsection{Harness for Agentic Execution}
\label{sec:related_harness}

Agent harness research studies the execution layer around a foundation model, covering tools, state, orchestration, validation, observability, and recovery. Recent surveys organize harness design around execution environment, tool interface, context management, lifecycle control, verification, and governance \cite{harness_survey,meng2026agent}. Agentic Harness Engineering evolves harness components through structured observability over files, trajectories, and decisions \cite{harness1}. Natural Language Agent Harnesses separate readable harness policy from runtime mechanisms \cite{harness2}. Meta Harness treats harness optimization as an outer loop over code, traces, and scores \cite{MetaHarness}. Beyond coding agents, harness-style designs have begun to appear in visual and embodied tasks. VASA maintains a persistent working mask for open ad hoc segmentation, turning visual progress into an inspectable state \cite{VASA}. SceneWeaver uses a reflective tool-based loop to synthesize 3D scenes under semantic and physical feedback \cite{SceneWeaver}. These systems indicate that explicit state, callable tools, validation gates, and shared memory can help complex layout text-to-image generation keep spatial intent readable, enforceable, and recoverable.

\section{Methodology}

\subsection{Overview}
\label{sec:framework_overview}

Complex 3D spatial text-to-image generation requires preserving object identity, relative position, visibility, camera configuration, and realism from language understanding to image synthesis. Existing controllable methods use layouts, 3D controls, layered generation, or agentic planning to improve spatial grounding \citep{st3d,SceneDesigner,Compass_control,Layercraft,li2025mccd}, but spatial requirements often remain scattered across prompts, conditions, and evaluation signals. Relations understood early may weaken after synthesis or disappear during revision. SpatialGuard addresses this by combining agent-based spatial reasoning with a Layout Harness, as shown in Fig.~\ref{fig:framework}.

Given an input prompt, SpatialGuard separates object mentions, spatial intentions, and appearance descriptions, then organizes them into a layout-centered generation loop. The system plans a structured scene, realizes it as visual evidence, checks whether the image follows the spatial intent, and repairs the layout when violations appear. In Fig.~\ref{fig:framework}, the input enters the Spatial Layout Architect, the Visual Realizer produces a candidate image, the Visual Alignment Critic performs structured validation, and failed items return through an executable repair plan.

The Layout Harness keeps this loop stable. Rule constraints make object, relation, scale, support, occlusion, visibility, and camera requirements checkable. Tool invocation maps revisions to operations such as moving objects, rescaling objects, and adjusting the camera. Shared knowledge stores layout decisions, validation records, and repair traces across rounds. Feedback loops send detected spatial errors back to planning rather than leaving them as passive scores. SpatialGuard therefore turns prompt-only sampling into a verifiable process that repeatedly plans, realizes, validates, and repairs the scene.

\subsection{SpatialGuard Framework}
\label{sec:spatialguard_framework}

SpatialGuard instantiates the common interface in Sec-.~\ref {sec:framework_overview} as an agentic layout generation framework. Prior layout-driven and 3D-controlled text-to-image methods make spatial control more explicit \citep{Layoutgpt,st3d, SceneDesigner, Compass_control}, but often pass the layout as a static generator condition. SpatialGuard instead treats the layout as an editable state that links language understanding, visual synthesis, and result validation, moving complex 3D spatial intent from implicit prompt attention into a state that can be planned, grounded, and inspected.

Following Sec.~\ref{sec:framework_overview}, $\bm{p}$ denotes the input prompt, $\bm{\mathcal{O}}=\{\bm{o}_{i}\}_{i=1}^{N}$ denotes the set of $N$ mentioned objects,$\bm{o}_{i}$ denotes the $i$th object. $\bm{\mathcal{C}}$ denotes the spatial constraint set extracted from $\bm{p}$, and $\bm{a}$ denotes the appearance and environment description. At round $t$, the layout state is written as
\begin{equation}
\bm{L}^{(t)}=
\left(
\left\{
\left(\bm{c}_{i},\bm{x}_{i}^{(t)},\bm{s}_{i}^{(t)},\bm{\alpha}_{i}^{(t)}\right)
\right\}_{i=1}^{N},
\bm{\kappa}^{(t)}
\right),
\end{equation}
\noindent where $\bm{c}_{i}$ is the category of object $\bm{o}_{i}$, $\bm{x}_{i}^{(t)}$ is the 3D position of object $\bm{o}_{i}$ at round $t$, $\bm{s}_{i}^{(t)}$ is its layout-readable size at round $t$, $\bm{\alpha}_{i}^{(t)}$ is its azimuth at round $t$, and $\bm{\kappa}^{(t)}$ is the camera parameter vector at round $t$. SpatialGuard operates on this state through three cooperative modules.

The \textbf{Spatial Layout Architect} converts the parsed spatial intent into an executable geometry carrier. It initializes the first layout and later revises the layout according to structured feedback:
\begin{equation}
\begin{gathered}
\bm{L}^{(0)}=\mathcal{A}_{s}\left(\bm{p},\bm{\mathcal{O}},\bm{\mathcal{C}},\bm{a}\right),\\
\bm{L}^{(t+1)}=\mathcal{A}_{s}\left(\bm{p},\bm{\mathcal{O}},\bm{\mathcal{C}},\bm{a},\bm{L}^{(t)},\bm{\Delta}^{(t)}\right),
\end{gathered}
\end{equation}
\noindent where $\mathcal{A}_{s}$ denotes the Spatial Layout Architect, $\bm{L}^{(0)}$ denotes the initial layout state, $\bm{L}^{(t+1)}$ denotes the next layout state after revision, and $\bm{\Delta}^{(t)}$ denotes the repair feedback produced by the Visual Alignment Critic at round $t$. The architect resolves object mentions, normalizes categories and counts, decomposes spatial phrases into 3D and screen space requirements, and assigns positions, sizes, orientations, and camera parameters. For example, a front right relation is represented through depth ordering and lateral placement, keeping it checkable after projection.

The \textbf{Visual Realizer} maps the current layout state into visual conditions and synthesizes a candidate image. Given $\bm{L}^{(t)}$, it renders the layout, derives instance-level masks, aligns object regions with the corresponding prompt tokens, and invokes the generator:
\begin{equation}
\bm{I}^{(t)}=
\mathcal{G}\left(
\bm{p},
\bm{a},
\rho\left(\bm{L}^{(t)}\right),
\psi\left(\bm{p},\bm{L}^{(t)}\right)
\right),
\end{equation}
\noindent where $\rho$ denotes the layout renderer, $\psi$ denotes the operator that produces instance masks and token-aligned object regions, $\mathcal{G}$ denotes the layout-conditioned image generator, and $\bm{I}^{(t)}$ denotes the candidate image generated in round $t$. The rendered layout provides global geometry, masks preserve object boundaries, and token alignment keeps appearance phrases attached to targets, transferring the planned 3D arrangement into pixel space.

The \textbf{Visual Alignment Critic} closes the loop by judging whether the candidate image respects the prompt and the current layout:
\begin{equation}
\left(\bm{v}^{(t)},\bm{\Delta}^{(t)}\right)=
\mathcal{A}_{c}\left(
\bm{p},
\bm{\mathcal{C}},
\bm{L}^{(t)},
\bm{I}^{(t)}
\right),
\end{equation}
\noindent where $\mathcal{A}_{c}$ denotes the Visual Alignment Critic, $\bm{v}^{(t)}$ is a structured validation record, and $\bm{\Delta}^{(t)}$ is the repair feedback for the next layout revision. The critic reads $\bm{\mathcal{C}}$, which contains the required object relations, support conditions, visibility requirements, and camera-related constraints extracted from $\bm{p}$. It localizes failures to concrete fields of $\bm{L}^{(t)}$, such as object position, size, azimuth, occlusion status, or camera configuration, instead of collapsing the judgment into a single scalar score. If $\bm{v}^{(t)}$ contains no failed item, SpatialGuard returns $\bm{I}^{(t)}$ as the final output. Otherwise, $\bm{\Delta}^{(t)}$ is passed back to the Spatial Layout Architect to produce $\bm{L}^{(t+1)}$.

This separation gives each module a clear responsibility: the Spatial Layout Architect makes spatial intent explicit, the Visual Realizer turns it into visual evidence, and the Visual Alignment Critic converts deviations into actionable layout feedback. SpatialGuard, therefore, supports closed-loop modeling from textual spatial intent to verifiable visual generation without requiring one-pass implicit spatial inference by the image generator.

\subsection{Layout Harness for Persistent Spatial Execution}
\label{sec:layout_harness}

Existing text-to-image systems make spatial control more explicit through layouts, 3D conditions, layered composition, and agent collaboration \citep{Layoutgpt,st3d, SceneDesigner, Compass_control, Layercraft,li2025mccd}. However, these signals are often used as local prompts, static conditions, or post hoc scores, which are fragile for relation-dense scenes. A relation parsed early may vanish during object repair, visibility adjustment, or camera change. The core challenge is therefore to preserve, retrieve, and enforce relations across planning, realization, validation, and revision.

Agent harness research provides an execution view for this challenge \citep{harness1,harness_survey,harness2}. Although a harness has no single mathematical definition, prior work commonly treats it as an external layer that organizes model calls, tools, state, validation, recovery, and stopping conditions. From this view, we distill four properties for complex spatial generation: rule constraints, tool invocation, shared knowledge, and feedback loops. SpatialGuard instantiates them as a Layout Harness around the editable layout state introduced in Sec-.~\ref {sec:spatialguard_framework}.

At refinement round $t$, where $t$ indexes the current iteration, the Layout Harness maintains
\begin{equation}
\bm{H}^{(t)}=
\left(
\bm{\mathcal{C}},
\bm{\mathcal{T}},
\bm{M}^{(t)}
\right).
\end{equation}
\noindent Here $\bm{H}^{(t)}$ is the harness state at round $t$, $\bm{\mathcal{C}}$ is the spatial constraint set extracted from the input prompt $\bm{p}$, $\bm{\mathcal{T}}$ is the executable layout tool library, and $\bm{M}^{(t)}$ is the shared knowledge state. $\bm{M}^{(t)}$ records the object manifest, parsed spatial predicates, previous layout states, validation records, and repair actions, giving all modules a durable reference beyond a single model call.

\begin{table*}[t]
\centering
\caption{Quantitative comparison of spatial layout faithfulness judged by three vision language models \cite{google2025gemini25pro,qwen2025qwen3vl,xai2025grok4}. Scores range from 1 to 10 and may be fractional, with all values reported to two decimal places. Higher scores are better. The best results are highlighted in bold and the second best results are underlined.}
\label{tab:llm_results}
\resizebox{\textwidth}{!}{
\begin{tabular}{@{}lcccccccc@{}}
\toprule
\textbf{Method} & \textbf{Presence $\uparrow$} & \textbf{Position $\uparrow$} & \textbf{Relation $\uparrow$} & \textbf{Depth $\uparrow$} & \textbf{Scale $\uparrow$} & \textbf{Support $\uparrow$} & \textbf{Framing $\uparrow$} & \textbf{Overall $\uparrow$} \\
\midrule
HunyuanImage-2.1 \cite{tencent2025hunyuanimage21} & \underline{8.59} & 7.98 & 7.84 & 7.25 & 7.79 & \underline{7.64} & 8.18 & \underline{7.90} \\
SD-3.5-L \cite{SD3.5} & 8.31 & \underline{8.08} & 7.70 & 7.34 & \underline{7.91} & 7.37 & \underline{8.46} & 7.88 \\
FLUX.1-dev \cite{blackforestlabs2024flux} & 8.13 & 7.62 & \underline{7.96} & \underline{7.49} & 7.57 & 7.18 & 7.92 & 7.70 \\
Self-Cross \cite{Self-cross} & 6.06 & 5.20 & 4.96 & 4.74 & 4.87 & 4.82 & 6.80 & 5.35 \\
T2I-R1 \cite{T2i-r1} & 6.49 & 5.91 & 6.62 & 6.25 & 5.58 & 5.39 & 6.68 & 6.13 \\
Janus-Pro-R1 \cite{Janus-pro-r1} & 6.83 & 6.25 & 6.55 & 6.18 & 6.04 & 5.66 & 6.55 & 6.29 \\
NextStep-1 \cite{NextStep-1} & 7.22 & 6.72 & 6.35 & 6.05 & 6.36 & 6.08 & 6.95 & 6.53 \\
\rowcolor{red!10}
\textbf{SpatialGuard (Ours)} & \textbf{9.26} & \textbf{8.83} & \textbf{9.38} & \textbf{9.67} & \textbf{9.39} & \textbf{9.45} & \textbf{9.58} & \textbf{9.37} \\
\bottomrule
\end{tabular}
}
\end{table*}

\noindent\textbf{Rule constraints}
make linguistic requirements inspectable. SpatialGuard stores relations such as front, left, support, or full visibility in $\bm{\mathcal{C}}$ and checks them against $\bm{L}^{(t)}$ and $\bm{I}^{(t)}$. As defined in Sec.~\ref{sec:spatialguard_framework}, $\bm{L}^{(t)}$ is the 3D layout state at round $t$, and $\bm{I}^{(t)}$ is the candidate image generated at round $t$. The Visual Alignment Critic produces $\bm{v}^{(t)}$ and $\bm{\Delta}^{(t)}$, where $\bm{v}^{(t)}$ lists passed and failed constraints and $\bm{\Delta}^{(t)}$ specifies the next layout revision.

\noindent\textbf{Tool invocation}
turns a revision instruction into an executable operation. Given the validation result, the harness selects and applies a sequence of tool actions:
\begin{equation}
\begin{gathered}
\bm{u}^{(t)}=\Gamma\left(\bm{v}^{(t)},\bm{\Delta}^{(t)},\bm{H}^{(t)}\right),\\
\tilde{\bm{L}}^{(t+1)}=\Omega\left(\bm{L}^{(t)},\bm{u}^{(t)}\right),\\
\bm{M}^{(t+1)}=\Xi\left(\bm{M}^{(t)},\tilde{\bm{L}}^{(t+1)},\bm{v}^{(t)},\bm{u}^{(t)}\right).
\end{gathered}
\end{equation}
\noindent In this equation, $\bm{u}^{(t)}$ is the selected sequence of tool actions at round $t$, $\Gamma$ is the tool selection function, $\Omega$ is the execution function that applies $\bm{u}^{(t)}$ to $\bm{L}^{(t)}$, $\tilde{\bm{L}}^{(t+1)}$ is the tool-executed intermediate layout state, $\Xi$ is the memory update function, and $\bm{M}^{(t+1)}$ is the updated shared knowledge state. The tools include object translation, object scaling, support adjustment, occlusion control, visibility restoration, focal length adjustment, and camera elevation adjustment. Because they operate on explicit layout fields, a failed relation can be repaired while preserving constraints that already pass validation.

\noindent\textbf{Shared knowledge}
keeps spatial decisions recoverable across rounds. The state $\bm{M}^{(t)}$ tells later modules which objects exist, which relations come from $\bm{p}$, which fields of $\bm{L}^{(t)}$ changed, and which failures remain. This avoids reliance on incomplete conversation history and makes each image traceable to a layout version, validation record, and repair action sequence.

\noindent\textbf{Feedback loops}
connect validation to the next execution step. When $\bm{v}^{(t)}$ contains failed constraints, $\bm{\Delta}^{(t)}$ is routed through $\Gamma$ to obtain $\bm{u}^{(t)}$, and $\Omega$ produces the tool-executed intermediate layout state $\tilde{\bm{L}}^{(t+1)}$, which is then used by the Spatial Layout Architect to form the next layout state $\bm{L}^{(t+1)}$. When no failed constraint remains, SpatialGuard returns $\bm{I}^{(t)}$ as the final image. Through this harness, text to image generation becomes a persistent spatial execution process in which language constraints are externalized, checked, repaired, and remembered until the layout and image satisfy the requested spatial intent.

\begin{figure*}[t]
  \centering
  \includegraphics[width=1.0\linewidth]{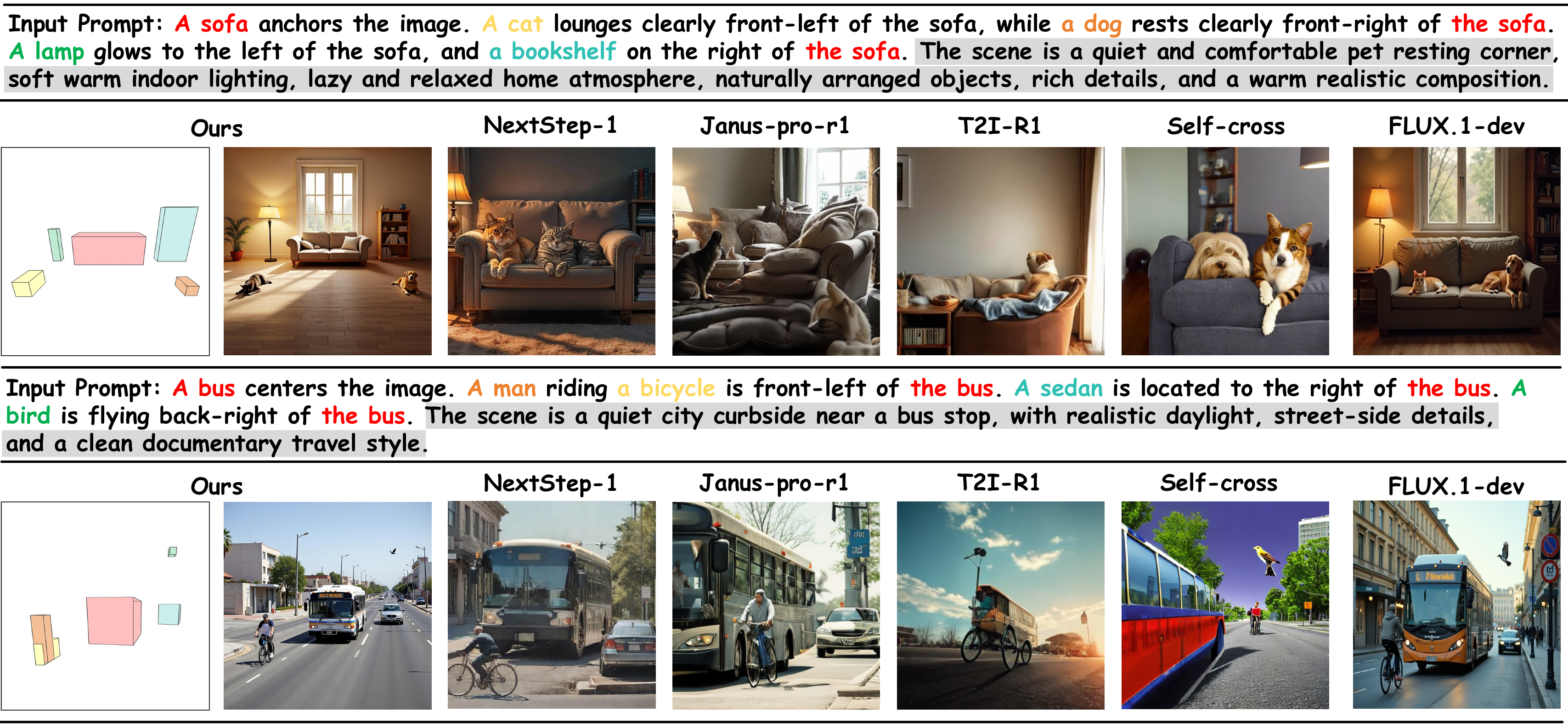}
  \caption{Qualitative comparison on complex spatial prompts. SpatialGuard preserves object completeness and relative layout more reliably than baselines, which often miss objects, truncate visible entities, or violate directional relations.
  }
\label{对比图}
\end{figure*}

\section{Experiments}
\subsection{Implementation Details}
SpatialGuard is a training-free framework, and all model weights remain frozen during evaluation. We conduct all experiments on a single NVIDIA H200 GPU with 143 GB of memory. Spatial Layout Architect and Visual Alignment Critic are implemented with GPT5 \cite{openai2025gpt5}, where the former produces executable 3D layouts and the latter returns structured validation records and repair instructions. Visual Realizer is built on FLUX.1 dev \cite{blackforestlabs2024flux}. We render layouts at 1024 resolution and use 512 resolution conditions for image synthesis. The default inference setting uses 25 denoising steps, a guidance scale of 3.5, and up to 4 layout verification and repair rounds.

\subsection{Quantitative Evaluation}

We evaluate spatial layout faithfulness with Gemini 2.5 Pro \cite{google2025gemini25pro}, Qwen3-VL \cite{qwen2025qwen3vl}, and Grok 4 \cite{xai2025grok4} using a fixed prompt set shared by all methods. The prompts cover object presence, directional relations, depth ordering, support, relative scale, and camera framing. We compare SpatialGuard with representative text to image methods \cite{tencent2025hunyuanimage21,SD3.5,Self-cross}. For each prompt, every method produces three final images with its recommended settings. Evaluators receive only the image and its prompt, without method identities or intermediate artifacts such as layout renderings, masks, validation records, or repair traces. This blind protocol uses identical prompts and instructions for all systems. No method is tuned on the evaluation prompts, and every comparison uses the same number of images. All three models score each image once, and the reported score for each dimension averages their judgments over all prompts and generated images.

Each image is evaluated on seven dimensions. Presence measures whether all requested objects and counts are visible and separable, with penalties for omissions, count errors, or severe fusion. Position measures how closely object locations and requested anchors match the prompt, while allowing partial credit for approximate placement. Relation evaluates atomic and compound spatial relations, with partial credit when only some components are satisfied. Depth covers front and behind order, occlusion, and residual visibility. Scale measures relative object sizes, distances between objects, and camera distance. Support evaluates contact with the specified surface and physical plausibility. Framing measures how well viewpoint, cropping, object visibility, and composition expose the requested layout. Overall is the arithmetic mean of these seven dimension scores. Appearance quality, style, and photorealism are ignored unless they prevent a reliable spatial judgment.

Evaluators may assign decimal scores on a 1 to 10 scale, and all reported values are rounded to two decimal places. Scores from 9 to 10 indicate clear and nearly complete satisfaction; 7 to 8 indicate a minor deviation; 5 to 6 indicate partial satisfaction or substantial ambiguity; 3 to 4 indicate major violations while relevant objects remain recognizable; and 1 to 2 indicate missing, contradictory, severely occluded, or unjudgeable evidence. Compound relations receive partial credit for visibly satisfied components. If a required object is absent, the associated relation receives a score from 1 to 2. Evaluators use only visible evidence and assign the lower score when evidence is ambiguous.

Table~\ref{tab:llm_results} shows that SpatialGuard obtains the highest Overall score of 9.37 and ranks first on all seven dimensions. HunyuanImage-2.1 \cite{tencent2025hunyuanimage21} is the strongest baseline at 7.90, narrowly ahead of SD-3.5-L \cite{SD3.5} at 7.88, leaving a 1.47 point gap to SpatialGuard. The largest gains over the best baseline for each dimension appear on Depth, Support, Scale, and Relation, at 2.18, 1.81, 1.48, and 1.42 points. Presence improves by 0.67, indicating that the main benefit lies in spatial structure rather than simple object inclusion. The Spatial Layout Architect converts language into explicit positions, depth order, orientations, and camera parameters, while the Visual Realizer transfers this plan into pixel space. The Visual Alignment Critic identifies visible violations, and the Layout Harness preserves satisfied constraints during targeted repair. Their coordination reduces relation reversals, inconsistent occlusion, and physically implausible layouts, which accounts for the consistent advantage across the spatial metrics. 

The smaller gain on Presence clarifies where the method helps most. Strong generators usually recover common objects, but they remain less reliable when several constraints must hold simultaneously. SpatialGuard instead preserves those constraints across planning, synthesis, inspection, and correction, yielding balanced performance.

\begin{table*}[t]
\centering
\caption{Ablation study using the same three vision language model evaluators \cite{google2025gemini25pro,qwen2025qwen3vl,xai2025grok4} and metrics reported in Table~\ref{tab:llm_results}. Scores range from 1 to 10, and higher scores are better. The best results are highlighted in bold and the second best results are underlined.}
\label{tab:消融}
\resizebox{\textwidth}{!}{
\begin{tabular}{@{}lcccccccc@{}}
\toprule
\textbf{Method} & \textbf{Presence $\uparrow$} & \textbf{Position $\uparrow$} & \textbf{Relation $\uparrow$} & \textbf{Depth $\uparrow$} & \textbf{Scale $\uparrow$} & \textbf{Support $\uparrow$} & \textbf{Framing $\uparrow$} & \textbf{Overall $\uparrow$} \\
\midrule
Ours w/o Spatial Layout Architect & \underline{9.01} & 8.10 & 8.62 & 8.70 & 8.52 & 8.40 & 9.00 & 8.62 \\
Ours w/o Visual Realizer & 8.74 & \underline{8.55} & 8.86 & 9.03 & 8.76 & \underline{8.95} & 8.84 & 8.82 \\
Ours w/o Visual Alignment Critic & 8.99 & 8.38 & \underline{9.08} & \underline{9.32} & 9.02 & 8.84 & \underline{9.31} & 8.99 \\
Ours w/o harness & 8.92 & 8.47 & 9.00 & 9.24 & \underline{9.18} & 9.03 & 9.22 & \underline{9.01} \\
\rowcolor{red!10}
\textbf{Ours(full)} & \textbf{9.26} & \textbf{8.83} & \textbf{9.38} & \textbf{9.67} & \textbf{9.39} & \textbf{9.45} & \textbf{9.58} & \textbf{9.37} \\
\bottomrule
\end{tabular}
}
\end{table*}

\subsection{Qualitative Evaluation}
\label{sec:qualitative_evaluation}

Figure \ref{对比图} compares SpatialGuard with recent text-to-image baselines on complex prompts containing object completeness, directional relations, and scene-level constraints. Baselines often produce visually plausible images, but their spatial structure is unstable. In the sofa scene, the prompt requires the cat and dog to appear in front of the sofa, while Self cross fails to preserve this layout. FLUX.1 dev \cite{blackforestlabs2024flux} generates only a partial bookshelf, and T2I-R1 \cite{T2i-r1} omits the lamp and bookshelf in the first example. In the bus scene, Janus-pro-r1 \cite{Janus-pro-r1} misses the bird, breaking the requested back right relation. SpatialGuard avoids these failures by maintaining an explicit layout state and checking object, relation, and visibility constraints before accepting the output. This leads to images that are not only realistic but also faithful to the full spatial intent.

Across both examples, the main difference is not isolated visual quality but whether several constraints survive together. SpatialGuard retains the required entities while keeping their relative positions readable after rendering. This is especially important for compound scenes, where correcting one object can disturb another relation or push an entity outside the frame. The explicit layout and validation loop localize such conflicts before acceptance, allowing the system to preserve completeness, direction, depth, and visibility within a composition.

\subsection{Ablation Studies}
\label{sec:ablation_studies}

Table~\ref{tab:消融} evaluates the four components with the same vision language model protocol \cite{google2025gemini25pro,qwen2025qwen3vl,xai2025grok4} as Table~\ref{tab:llm_results}. The complete system reaches an Overall score of 9.37. The second best score changes by criterion: the Architect ablation leads Presence at 9.01, the Realizer ablation leads Position and Support at 8.55 and 8.95, the Critic ablation leads Relation, Depth, and Framing at 9.08, 9.32, and 9.31, and the Harness ablation leads Overall at 9.01. This distribution follows the roles of the components. Without the Architect, Presence remains high, but Position and Relation fall to 8.10 and 8.62, showing that precise geometry needs an explicit layout plan. The Realizer transfers this plan into image evidence, while the Critic detects visible violations for repair. The Harness preserves decisions across rounds, affecting Scale, Support, and Overall. The gap from the full configuration shows that SpatialGuard benefits from their interaction rather than one isolated module.

\section{Conclusion}
\label{sec:conclusion}

This paper introduces SpatialGuard, a structured layout-guided framework for complex 3D spatial text-to-image generation. SpatialGuard externalizes spatial intent into an editable 3D layout state and organizes generation as planning, realization, validation, and repair, rather than leaving spatial relations to implicit prompt interpretation. Its Layout Harness stabilizes this process with rule constraints, tool invocation, shared knowledge, and feedback loops, keeping object relations, visibility, and camera constraints recoverable across iterations. Experiments show clear gains in spatial pose, relation grounding, and measurement accuracy over strong text-to-image baselines. Qualitative and ablation results further show these gains come from verifiable layout execution.

\section{Acknowledgements}
This work is supported by the 2026 AI Breakthrough Initiative Research Project 2026JLGJ0001GX.

\section*{Limitations}

SpatialGuard focuses on spatially grounded scenes that can be expressed through object lists, relations, visibility constraints, and camera parameters. Prompts with highly abstract artistic intent or intentionally ambiguous spatial descriptions may require additional interpretation rules. Since the framework performs planning, validation, and repair before producing the final image, its inference cost is higher than a single direct call to a text-to-image model. In addition, the final visual quality still depends on the capability of the underlying image generator. Future work can extend the tool library to finer physical interactions and improve efficiency through lighter validation strategies.

\bibliography{custom}

\end{document}